\documentclass[journal]{IEEEtran}
\usepackage{amsmath,amsfonts}
\usepackage{algorithmic}
\usepackage{algorithm}
\usepackage{array}
\usepackage[caption=false,font=normalsize,labelfont=sf,textfont=sf]{subfig}
\usepackage[table]{xcolor}

\usepackage{textcomp}
\usepackage{stfloats}
\usepackage{url}
\usepackage{verbatim}
\usepackage{graphicx}
\usepackage{cite}
\usepackage{hyperref}
\begin{document}
\definecolor{lightgray}{RGB}{240,240,240}
\title{Learning Regional Snow Water Equivalent and Snow Height Variations from Sentinel-1 InSAR Acquisitions}
\author{Luca~Barco~$^{1,2 *}$,~\IEEEmembership{Member,~IEEE,}
        Lorenzo~Innocenti~$^{1,2 *}$,~\IEEEmembership{Member,~IEEE,}
        Bianca~Bartoli~$^{1}$,
        Claudio~Rossi~$^{2}$,
        Edoardo~Arnaudo~$^{2}$
        and Paolo~Garza~$^{1}$
\thanks{$^{*}$Equal contribution.}
\thanks{$^{1}$Politecnico di Torino, Dipartimento di Automatica e Informatica, Torino, Italy. \textit{name.surname@polito.it}}
\thanks{$^{2}$Fondazione LINKS, AI Data and Space, Italy. \textit{name.surname@linksfoundation.com}}
\thanks{© 20XX IEEE. Personal use of this material is permitted. Permission from IEEE must be obtained for all other uses, in any current or future media, including reprinting/republishing this material for advertising or promotional purposes, creating new collective works, for resale or redistribution to servers or lists, or reuse of any copyrighted component of this work in other works.This work was carried out in the context of Space IT Up project funded by the Italian Space Agency (ASI) and the Ministry of University and Research (MUR) under contract n. 2024-5-E.0 - CUP n. I53D2400060005 and NODES project through the MUR—M4C2 1.5 of PNRR under Grant ECS00000036}
}

\markboth{IEEE GEOSCIENCE AND REMOTE SENSING LETTERS, VOL. XX, YYYY}%
{Shell \MakeLowercase{\textit{et al.}}: A Sample Article Using IEEEtran.cls for IEEE Journals}


\maketitle
\begin{abstract}
Managing water resources in mountainous regions depends heavily on reliable Snow Water Equivalent (SWE) and Snow Height (HS) data, yet these variables remain difficult to track at scale. This study evaluates three machine learning architectures (XGBoost, U-Net and SegFormer) for the joint estimation of SWE and HS variations from Sentinel-1 InSAR data over the Italian Alps, using the IT-SNOW reanalysis as reference. SegFormer achieves the best results on both targets, with an MAE of 10.391 cm for HS and 27.113 mm w.e. for SWE and the lowest variability across initializations. A feature sensitivity analysis shows that including all available features does not guarantee the lowest error, with model- and task-specific sensitivities. 
Spatial metrics (\textsuperscript{2}, Pearson's r) separate the three architectures far more clearly than mean error (MAE, RMSE) does, and decomposing the error per window attributes most of it to a systematic offset in the estimated mean variation rather than to the spatial pattern.
\end{abstract}


\section{Introduction}
\label{sec:intro}
\IEEEPARstart{E}stimating SWE and HS in mountainous regions is essential for water resource management, hydrological forecasting, and flood risk assessment. While ground-based measurements provide high accuracy at point locations, their sparse spatial coverage limits operational snow monitoring over large alpine domains. Synthetic Aperture Radar (SAR) offers a scalable alternative due to its all-weather capabilities and sensitivity to snowpack properties, with the Sentinel-1 C-band constellation becoming central for its high resolution, temporal continuity, and free availability, as noted by Jans et al. (2025)~\cite{JANS2025114477}.
Two families of retrieval methods have emerged. Interferometric SAR (InSAR) exploits the phase delay caused by signal refraction within the snowpack to estimate SWE variations~\cite{Oveisgharan, 10.3389/frsen.2024.1481848}. However, Sentinel-1's 6-day repeat cycle is susceptible to temporal decorrelation, often limiting this approach to dry, stable snow; Palomaki and Sproles (2023) showed that L-band sensors mitigate these effects through longer wavelengths that maintain higher coherence, though they lack Sentinel-1's temporal accessibility~\cite{PALOMAKI2023113744}. Amplitude-based methods instead exploit the ratio of cross-polarized to co-polarized backscatter, as developed by Lievens et al. (2019) to estimate HS in the Northern Hemisphere mountains~\cite{Lievens2019}. The global transferability of this C-band volume scattering signal remains under investigation: Hoppinen et al. (2024) observed significant degradation (RMSE of 0.92~m) in forested areas and deep snowpacks across NASA SnowEx sites~\cite{tc-18-5407-2024}, suggesting a signal response highly sensitive to local stratigraphy and vegetation.
Machine learning (ML) addresses the non-linear relationships between radar signals and snow properties. Dunmire et al. (2024) proposed a physics-informed XGBoost framework over the Alps, combining Sentinel-1 polarimetric features with meteorological forcing to downscale atmospheric reanalysis to $100$~m snow depth maps~\cite{DUNMIRE2024114369}, while Betato et al. (2025) introduced \textit{MAPunet}~\cite{BETATO2025101477}, a U-Net-based approach~\cite{10.1007/978-3-319-24574-4_28} for pixel-wise snow depth regression over Davos, Switzerland. Both estimate absolute quantities at finer resolution over smaller domains, relying primarily on backscatter and meteorological inputs rather than on aggregated interferometric features. Three questions therefore remain open. First, most approaches focus on single-target estimation, either SWE or HS, despite the physical coupling between these parameters. Second, the optimal combination of InSAR-derived features (coherence, intensity, phase, displacement) remains unclear, particularly when integrating auxiliary geophysical data. Third, how different ML paradigms exploit the same set of aggregated InSAR observables has not been systematically compared.
Complementary to these retrievals, gridded reanalysis products have emerged as valuable resources for large-scale algorithm development. Avanzi et al. (2023) presented IT-SNOW~\cite{essd-15-639-2023}, a reanalysis for Italy that blends numerical modeling, \textit{in situ} observations, and satellite data at $500$~m resolution, which we adopt as ground truth throughout.
This work compares three ML architectures, i.e., XGBoost, U-Net with ResNet backbone, and SegFormer, for the simultaneous retrieval of SWE and HS variations from Sentinel-1 InSAR data over the Italian Alps. The main contributions are: (1) formulation of snow estimation as a joint multi-target regression problem, enabling deep learning models to predict both targets concurrently; (2) a feature sensitivity analysis revealing that including all available features does not guarantee optimal performance, with model- and task-specific sensitivities, complemented by a group-wise ablation quantifying the contribution of the interferometric channels as a whole; and (3) an assessment showing that the three paradigms are separated far more clearly by spatial correlation metrics than by mean error, with SegFormer attaining the lowest error and variability across initializations. We train on two full years (09 January 2021-30 December 2022) and test on an independent period (30 December 2022- 24 March 2023).
Source code and dataset are publicly available at \url{https://github.com/links-ads/insar-regional-snow-mapping} and \url{https://huggingface.co/datasets/links-ads/insar-regional-snow-mapping}, respectively.

\section{Dataset}
\label{sec:dataset}

\begin{figure}
    \begin{minipage}[b]{\linewidth}
      \centerline{\includegraphics[width=\linewidth]{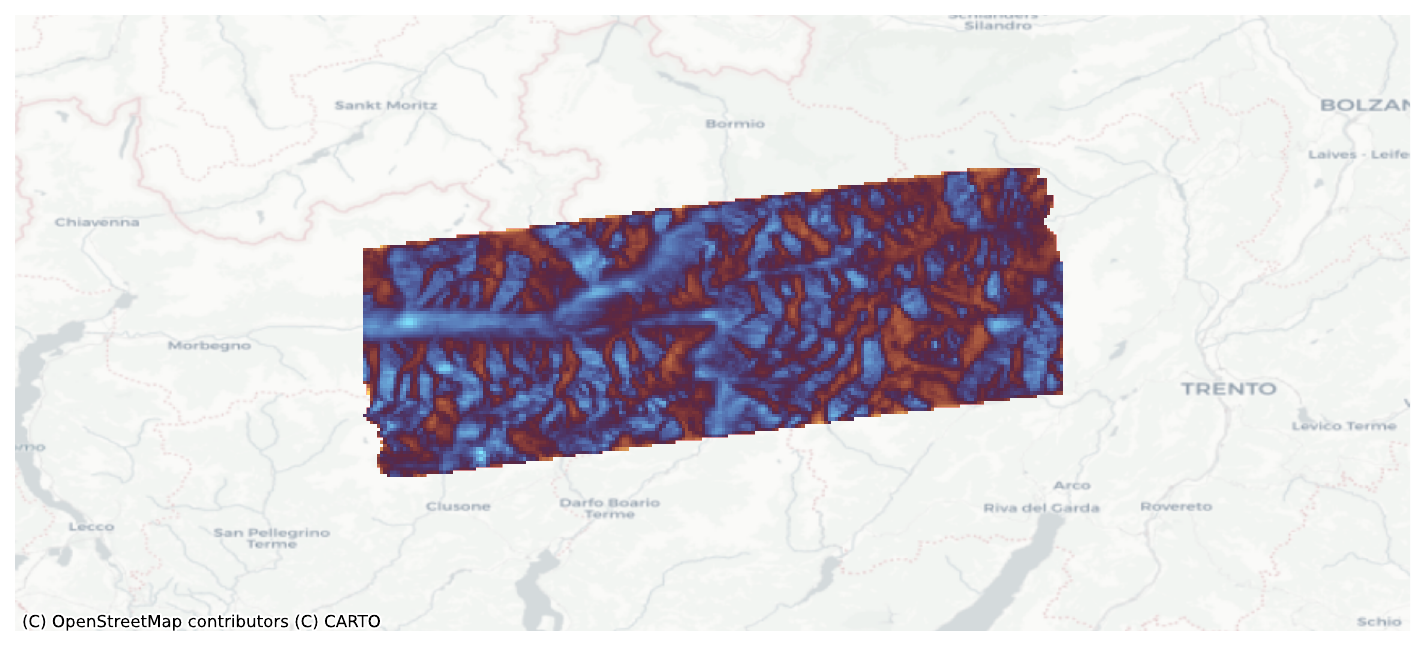}}
    %  \vspace{1.5cm}
      \caption{Example of InSAR Intensity Interferogram over the Area of Interest  (Orobie Alps and Adamello)}\medskip
      \label{img:aoi}
    \end{minipage}

\end{figure}

The case study area considered in this work covers the Orobie Alps and Adamello ranges in Italy (Figure \ref{img:aoi}). Their varied terrain and snow accumulation patterns make them representative of complex alpine environments and suitable for testing retrieval algorithms designed for broader mountain regions.

The InSAR maps were produced from Sentinel-1A Single Look Complex (SLC)\footnote{\url{https://documentation.dataspace.copernicus.eu/Data/SentinelMissions/Sentinel1.html}} images acquired in Interferometric Wide (IW) swath mode using TOPSAR (VV polarization, 12-day repeat cycle), along ascending relative orbit 117, sub-swath IW1, bursts 4-5 of slice 11.
InSAR processing was performed using ESA SNAP\footnote{\url{https://earth.esa.int/eogateway/tools/snap}} following the reference S1TBX TOPSAR interferometry workflow\footnote{\url{https://step.esa.int/docs/tutorials/S1TBX\%20TOPSAR\%20Interferometry\%20with\%20Sentinel-1\%20Tutorial_v2.pdf}}: TOPSAR split, orbit file application, back-geocoding coregistration refined by Enhanced Spectral Diversity, interferogram formation, coherence estimation, TOPS debursting, multilooking, Goldstein filtering (coherence threshold 0.2), phase unwrapping, displacement computation via line-of-sight geometry, and terrain correction using Copernicus DEM\footnote{\url{https://dataspace.copernicus.eu/explore-data/data-collections/copernicus-contributing-missions/collections-description/COP-DEM}} (30m). Multilooking used a single azimuth look and a range look factor selected per scene to yield approximately square ground pixels. Unwrapping was performed with SNAPHU\footnote{\url{https://web.stanford.edu/group/radar/softwareandlinks/sw/snaphu}} in deformation cost mode, initialized by minimum spanning tree, over a $10 \times 10$ tile grid with 200-pixel overlap and a tile cost threshold of 500. Pixels below the coherence threshold, and those for which unwrapping does not converge, are propagated as missing values.
Seven per-pixel features were extracted: five interferometric quantities, i.e. (1) coherence, (2) intensity [dB], (3-4) wrapped/unwrapped phase [rad], and (5) displacement [m], and two geometric variables derived from the acquisition geometry and the DEM, i.e. (6) local incidence angle [deg] and (7) elevation [m].
Due to C-band limitations in wet snow conditions, Sentinel-3 SLSTR Land Surface Temperature (LST)\footnote{\url{https://documentation.dataspace.copernicus.eu/Data/SentinelMissions/Sentinel3.html}} was acquired to help in discriminate dry/wet snow regimes. For each interferometric pair, two thermal features were computed: (8) mean LST [K] and (9) Positive Degree Days [°C] (PDD, cumulative degrees above 0°C between acquisition dates).
Regarding temporal information, (10-13) sine and cosine encodings of start and end acquisition dates were included to capture seasonal patterns.

Ground truth was derived from IT-SNOW\cite{essd-15-639-2023}, a 500m gridded reanalysis for the Italian Alps providing daily HS [cm] and SWE [mm w.e.] through assimilation of in-situ measurements and model outputs, as instantaneous snapshots at 11:00 UTC taken as representative of the day. The dataset was spatially cropped to the Area of Interest, excluding areas outside Italian borders where IT-SNOW is not defined. Target labels were computed as temporal differences: $\Delta$HS and $\Delta$SWE between consecutive SAR acquisition dates. 
The final dataset spans 09 January 2021 through 24 March 2023, comprising 64 temporal samples. All data sources were resampled to 500m per pixel resolution using bilinear interpolation to align with the IT-SNOW spatial resolution.

\section{Methodology}
\label{sec:methodology}
This study employs a comparative analysis of three ML architectures: XGBoost \cite{10.1145/2939672.2939785}, U-Net \cite{10.1007/978-3-319-24574-4_28} with ResNet \cite{7780459} backbone, and SegFormer \cite{xie2021SegFormer}. These methods represent three distinct machine‑learning paradigms: classical gradient boosting, convolutional encoder–decoder networks, and vision transformers.

\subsection{Problem Formulation}

The SWE and HS estimation task is framed as a multi‑target supervised regression problem focused on temporal changes. Since the IT‑SNOW reanalysis provides daily absolute values, while InSAR measures only differences between acquisition dates, the prediction targets are expressed as \mbox{$\Delta\text{SWE}(t_2, t_1) = \text{SWE}(t_2) - \text{SWE}(t_1)$} and \mbox{$\Delta\text{HS}(t_2, t_1) = \text{HS}(t_2) - \text{HS}(t_1)$}
, i.e., their variations between two SAR acquisitions, where $t_1$ and $t_2$ represent the dates of the reference and secondary SAR images, respectively. 

Given a set of input features $\mathbf{X} \in \mathbb{R}^{H \times W \times C}$ derived from InSAR processing, LST and temporal representation, where $H$ and $W$ denote the spatial dimensions and $C$ represents the number of feature channels, the objective is to predict the spatially distributed snow change maps $\Delta\mathbf{Y} \in \mathbb{R}^{H \times W}$. 

\subsection{Model Architectures}
The first model considered in this study is \textbf{XGBoost} \cite{10.1145/2939672.2939785}, which builds an ensemble of decision trees sequentially, with each tree trained to reduce the residual errors of the previous ones. The model operates on pixel-wise feature vectors without explicitly leveraging spatial relationships between neighboring pixels. In the context of this work, the results for SWE and HS from XGBoost refers to two different model, one trained on each target.

\textbf{U-Net} \cite{10.1007/978-3-319-24574-4_28} is a convolutional architecture consisting of a contracting path (encoder) that captures contextual information through successive convolutions and pooling operations, and an expansive path (decoder) that enables precise localization through upsampling and skip connections.
In this work, the encoder is replaced with a ResNet-18 backbone \cite{7780459}, which provides more robust feature extraction through residual connections.
The decoder path combines high-resolution features from the encoder (via skip connections) with upsampled feature maps to progressively recover spatial resolution. The final layer produces a two-channel output representing the predicted HS and SWE maps.

\textbf{SegFormer} \cite{xie2021SegFormer} employs a hierarchical Transformer encoder together with a lightweight MLP decoder. Unlike CNNs, Transformers operate on image patches using self-attention, enabling the capture of long-range dependencies and global context. Specifically, the MiT-B0 encoder has been considered, which provides an efficient multi-stage architecture with progressive downsampling and multi-scale feature extraction. The MLP decoder then aggregates these features to produce the final HS and SWE maps.

Model performance is evaluated using Mean Absolute Error (MAE), Root Mean Squared Error (RMSE), the coefficient of determination R\textsuperscript{2} and Pearson's correlation coefficient $r$.

\section{Experiments}
\label{sec:exps}
 % \begin{figure*}
%     \centering
%     \includegraphics[width=\linewidth]{images/all_plots.pdf} 
%     \caption{Average and standard deviation of ablation results.}
%     \label{fig:plots}
% \end{figure*}

\subsection{Implementation details}
All models were trained on data from 09 January 2021 to 30 December 2022 using 48 samples for training and 9 for validation, the latter sampled at approximately two-month intervals across the two seasons so as to cover all snow regimes rather than a single contiguous period. The final 7 temporal windows (30 December 2022 - 24 March 2023) were held out as an independent test set. Each experiment was repeated with three random initializations (seeds: 42, 123, 2024).
XGBoost operated on pixel-wise feature vectors, treating each pixel independently, with two separate models for HS and SWE, pseudo-Huber objective, 500 estimators and learning rate of 0.1; pixels with missing feature values are handled natively by the tree splits, which learn a default direction for each split. U-Net and SegFormer performed joint multi-target regression on full patches, so missing feature values are imputed as zero, and share the same configuration: spatially contiguous patches of $256 \times 256$, AdamW optimizer (lr = $10^{-5}$, weight decay = $10^{-2}$), cosine annealing scheduler, batch size of 16, maximum of 200 epochs with early stopping patience of 70, and a Huber loss ($\delta=5$) masked to the pixels where the IT-SNOW reference is defined; SegFormer additionally employed decoder dropout of 0.2 and encoder drop path rate of 0.1. 
The feature sensitivity analysis was conducted by excluding each feature in turn and measuring test set performance. Thermal and temporal encodings were retained throughout, as they are essential for capturing seasonal patterns and distinguishing dry from wet regimes. The configurations with the lowest error were then used for a single-target comparison, and a further configuration was trained with all five interferometric bands removed.
% ===== Tabelle I-III: una colonna, impilate =====
\begin{table}[t!]
\centering
\resizebox{\columnwidth}{!}{%
\begin{tabular}{|l|cc|cc|}
\hline
 & \multicolumn{2}{c|}{\textbf{HS}} & \multicolumn{2}{c|}{\textbf{SWE}} \\
 & \textbf{MAE} & \textbf{RMSE} & \textbf{MAE} & \textbf{RMSE} \\ \hline
\rowcolor{lightgray} \textbf{All Features} & $14.429 \pm 1.164$ & $20.076 \pm 2.941$ & $31.238 \pm 0.682$ & $43.366 \pm 0.736$ \\ \hline
\multicolumn{5}{|l|}{\textbf{Excluded Feature}} \\ \hline
Coherence & $12.945 \pm 0.641$ & $17.080 \pm 0.544$ & $\mathbf{29.597 \pm 1.450}$ & $\mathbf{42.322 \pm 1.679}$ \\
Intensity & $12.575 \pm 0.123$ & $16.654 \pm 0.098$ & $30.474 \pm 1.639$ & $43.389 \pm 1.271$ \\
Phase & $\mathbf{12.374 \pm 0.294}$ & $\mathbf{16.465 \pm 0.392}$ & $31.420 \pm 1.554$ & $43.947 \pm 1.217$ \\
Unwrapped phase & $12.890 \pm 0.632$ & $17.207 \pm 0.758$ & $29.960 \pm 0.916$ & $42.831 \pm 1.022$ \\
Incidence angle & $12.423 \pm 0.685$ & $16.530 \pm 0.639$ & $30.306 \pm 1.241$ & $42.832 \pm 0.997$ \\
Elevation & $12.770 \pm 0.379$ & $16.513 \pm 0.380$ & $31.307 \pm 1.442$ & $43.338 \pm 1.844$ \\
Displacement & $13.389 \pm 0.619$ & $17.496 \pm 0.523$ & $29.893 \pm 0.506$ & $42.439 \pm 0.669$ \\ \hline
\end{tabular}%
}
\vspace{2mm}
\caption{Feature sensitivity results for XGBoost}
\label{tab:ablation_xgboost}

\resizebox{\columnwidth}{!}{%
\begin{tabular}{|l|cc|cc|}
\hline
 & \multicolumn{2}{c|}{\textbf{HS}} & \multicolumn{2}{c|}{\textbf{SWE}} \\
 & \textbf{MAE} & \textbf{RMSE} & \textbf{MAE} & \textbf{RMSE} \\ \hline
\rowcolor{lightgray} \textbf{All Features} & $12.754 \pm 0.353$ & $18.224 \pm 0.580$ & $31.635 \pm 1.670$ & $45.263 \pm 1.251$ \\ \hline
\multicolumn{5}{|l|}{\textbf{Excluded Feature}} \\ \hline
Coherence & $11.919 \pm 1.553$ & $16.542 \pm 2.176$ & $31.589 \pm 3.950$ & $44.718 \pm 4.481$ \\
Intensity & $13.362 \pm 1.700$ & $19.097 \pm 3.154$ & $31.959 \pm 3.872$ & $45.507 \pm 6.094$ \\
Phase & $12.624 \pm 1.323$ & $18.209 \pm 1.899$ & $30.715 \pm 2.647$ & $44.715 \pm 3.374$ \\
Unwrapped phase & $12.207 \pm 1.433$ & $17.537 \pm 2.110$ & $29.949 \pm 2.874$ & $43.851 \pm 3.817$ \\
Incidence angle & $11.871 \pm 0.922$ & $16.877 \pm 0.877$ & $29.272 \pm 1.591$ & $42.619 \pm 1.704$ \\
Elevation & $12.770 \pm 1.667$ & $17.458 \pm 1.512$ & $32.310 \pm 4.325$ & $45.340 \pm 3.715$ \\
Displacement & $\mathbf{11.464 \pm 0.262}$ & $\mathbf{16.142 \pm 0.418}$ & $\mathbf{28.032 \pm 0.625}$ & $\mathbf{41.422 \pm 1.022}$ \\ \hline
\end{tabular}%
}
\vspace{2mm}
\caption{Feature sensitivity results for U-Net}
\label{tab:ablation_unet}

\resizebox{\columnwidth}{!}{%
\begin{tabular}{|l|cc|cc|}
\hline
 & \multicolumn{2}{c|}{\textbf{HS}} & \multicolumn{2}{c|}{\textbf{SWE}} \\
 & \textbf{MAE} & \textbf{RMSE} & \textbf{MAE} & \textbf{RMSE} \\ \hline
\rowcolor{lightgray} \textbf{All Features} & $10.391 \pm 0.445$ & $14.684 \pm 0.210$ & $27.113 \pm 1.227$ & $40.388 \pm 1.254$ \\ \hline
\multicolumn{5}{|l|}{\textbf{Excluded Feature}} \\ \hline
Coherence & $11.167 \pm 0.514$ & $15.188 \pm 0.378$ & $29.012 \pm 0.925$ & $41.130 \pm 0.508$ \\
Intensity & $10.598 \pm 1.607$ & $14.955 \pm 1.847$ & $27.354 \pm 3.309$ & $40.578 \pm 3.763$ \\
Phase & $11.940 \pm 0.963$ & $16.502 \pm 1.594$ & $29.435 \pm 0.925$ & $42.695 \pm 0.833$ \\
Unwrapped phase & $9.799 \pm 3.749$ & $13.697 \pm 5.118$ & $\mathbf{25.673 \pm 0.963}$ & $\mathbf{38.594 \pm 0.712}$ \\
Incidence angle & $10.336 \pm 0.771$ & $14.737 \pm 1.089$ & $27.053 \pm 1.784$ & $40.269 \pm 2.425$ \\
Elevation & $10.070 \pm 0.740$ & $14.103 \pm 0.627$ & $25.790 \pm 1.003$ & $38.717 \pm 0.742$ \\
Displacement & $\mathbf{9.856 \pm 0.783}$ & $\mathbf{14.089 \pm 0.756}$ & $26.295 \pm 1.685$ & $39.316 \pm 1.758$ \\ \hline
\end{tabular}%
}
\vspace{2mm}

\caption{Feature sensitivity results for SegFormer}
\label{tab:ablation_segformer}

\end{table}

% ===== Tabella IV: float separato a piena larghezza =====
\begin{table*}[t!]
\centering
\resizebox{\textwidth}{!}{%
\begin{tabular}{|l|l|cccc|cccc|}
\hline
 & & \multicolumn{4}{c|}{\textbf{HS}} & \multicolumn{4}{c|}{\textbf{SWE}} \\
\textbf{Model} & \textbf{Configuration} & \textbf{MAE} & \textbf{RMSE} & \textbf{R\textsuperscript{2}} & \textbf{r} & \textbf{MAE} & \textbf{RMSE} & \textbf{R\textsuperscript{2}} & \textbf{r} \\ \hline\hline
\multicolumn{10}{|l|}{\textbf{(a) Baseline --- all features, multi-target}} \\ \hline
\rowcolor{lightgray} XGBoost & -- & $14.429 \pm 1.164$ & $20.076 \pm 2.941$ & $-2.631 \pm 0.697$ & $0.002 \pm 0.006$ & $31.238 \pm 0.682$ & $43.366 \pm 0.736$ & $-1.468 \pm 0.152$ & $0.097 \pm 0.019$ \\
\rowcolor{lightgray} U-Net & -- & $12.754 \pm 0.353$ & $18.224 \pm 0.580$ & $-2.138 \pm 0.169$ & $0.018 \pm 0.028$ & $31.635 \pm 1.670$ & $45.263 \pm 1.251$ & $-1.698 \pm 0.310$ & $0.081 \pm 0.034$ \\
\rowcolor{lightgray} SegFormer & -- & $10.391 \pm 0.445$ & $14.684 \pm 0.210$ & $-0.800 \pm 0.082$ & $0.065 \pm 0.035$ & $27.113 \pm 1.227$ & $40.388 \pm 1.254$ & $-0.667 \pm 0.069$ & $0.139 \pm 0.004$ \\ \hline\hline
\multicolumn{10}{|l|}{\textbf{(b) Multi-target vs.\ single-target at matched feature sets}} \\ \hline
U-Net & Multi, no displ. & $11.464 \pm 0.262$ & $16.142 \pm 0.418$ & $-1.342 \pm 0.254$ & $0.056 \pm 0.069$ & $28.032 \pm 0.625$ & $41.422 \pm 1.022$ & $-0.956 \pm 0.285$ & $0.141 \pm 0.067$ \\
U-Net & Single, no displ. & $12.223 \pm 1.701$ & $17.013 \pm 1.745$ & $-1.636 \pm 0.619$ & $0.024 \pm 0.010$ & $30.119 \pm 1.702$ & $42.929 \pm 0.626$ & $-1.190 \pm 0.080$ & $0.089 \pm 0.005$ \\ \hline
SegFormer & Multi, no displ. & $\mathbf{9.856 \pm 0.783}$ & $14.089 \pm 0.756$ & $-0.633 \pm 0.158$ & $0.061 \pm 0.064$ & $26.295 \pm 1.685$ & $39.316 \pm 1.758$ & $-0.558 \pm 0.169$ & $0.135 \pm 0.045$ \\
SegFormer & Single, no displ. & $10.674 \pm 1.322$ & $14.778 \pm 1.351$ & $-0.878 \pm 0.387$ & $0.056 \pm 0.049$ & -- & -- & -- & -- \\ \hline
SegFormer & Multi, no unwr.\ ph. & $9.890 \pm 0.211$ & $\mathbf{13.991 \pm 0.181}$ & $\mathbf{-0.613 \pm 0.031}$ & $0.060 \pm 0.021$ & $\mathbf{25.673 \pm 0.963}$ & $\mathbf{38.594 \pm 0.712}$ & $\mathbf{-0.442 \pm 0.080}$ & $0.111 \pm 0.024$ \\
SegFormer & Single, no unwr.\ ph. & -- & -- & -- & -- & $28.541 \pm 2.470$ & $41.200 \pm 3.365$ & $-0.927 \pm 0.403$ & $0.113 \pm 0.036$ \\ \hline
\end{tabular}%
}
\vspace{2mm}
\caption{Spatial metrics for the three architectures and comparison between multi-target and single-target training. Block (a) uses the complete feature set; block (b) uses the configuration with the lowest error in Tables~\ref{tab:ablation_unet} and \ref{tab:ablation_segformer}. }
\label{tab:single_target}
\end{table*}

\subsection{Results}

All metrics are computed per test window and averaged over the seven windows; RMSE is the root of the mean squared error over the same windows. Results are reported as mean $\pm$ std across the three random seeds.

The feature sensitivity results in Tables~\ref{tab:ablation_xgboost}, \ref{tab:ablation_unet} and \ref{tab:ablation_segformer} show that individual input features contribute differently to each architecture, and that using all available features does not guarantee the lowest error for any of the three models. For SegFormer on HS, excluding unwrapped phase achieves the lowest mean error (9.799 cm) but with high instability across seeds (std of 3.749 cm), whereas displacement exclusion offers a better balance between accuracy (9.856 cm) and stability (std of 0.783 cm). For SWE, unwrapped phase and elevation exclusion give the lowest errors (25.673 and 25.790 mm w.e.). U-Net shows a similar but weaker pattern, with displacement exclusion yielding the lowest error for both HS (11.464 cm) and SWE (28.032 mm w.e.). 

XGBoost behaves in the opposite way. Every single-feature exclusion improves on the all-features configuration for both targets, and by a wide margin on HS, where the MAE drops from 14.429 to between 12.374 and 13.389 cm. The complete configuration is also the least stable across seeds, with a standard deviation of 2.941 cm in RMSE for HS against 0.098 to 0.758 cm for all reduced sets. This points to redundancy among the aggregated features: operating pixel-wise, XGBoost has no spatial context with which to disambiguate correlated inputs. Phase exclusion yields its lowest HS error (12.374 cm) and coherence exclusion its lowest SWE error (29.597 mm w.e.), confirming that the informative subset is task-dependent, yet even in its best configuration XGBoost yields higher errors than SegFormer on both targets.
Table~\ref{tab:single_target} adds R\textsuperscript{2}, that normalises the error by the target variance of each test window, and Pearson's $r$, that measures the agreement of spatial patterns independently of offset and amplitude. Block (a) reports the three architectures with the complete feature set, and block (b) compares multi-target with single-target training at matched feature sets, since single-target models were trained only with the exclusion that performed best for the corresponding target. R\textsuperscript{2} is negative for every configuration and both targets, ranging from $-0.442$ to $-2.631$, and $r$ never exceeds 0.141. Per window, the normalised error decomposes into a squared mean offset and a dispersion term, $-R^2 = (\Delta\mu/\sigma_r)^2 + (\sigma_p/\sigma_r)^2 - 2r(\sigma_p/\sigma_r)$, where $\Delta\mu$ is the difference between predicted and reference window means and $\sigma_p$, $\sigma_r$ their spatial standard deviations. The offset term accounts for most of the negative values: 0.509 of $-0.800$ for SegFormer on HS (64\%), 1.383 of $-2.138$ for U-Net (65\%) and 1.662 of $-2.631$ for XGBoost (63\%), with comparable proportions on SWE. The remainder reflects the combination of under-dispersion and near-zero pattern correlation. The dominant error is therefore a systematic misestimation of the window-mean variation rather than of its spatial distribution, and it is in principle correctable by calibration, whereas the low $r$ values indicate that the spatial component of the signal is largely not recovered.
Single-target training degrades MAE, RMSE and R\textsuperscript{2} in all four comparisons of block (b), with SegFormer moving from 9.856 to 10.674 cm on HS and from 25.673 to 28.541 mm w.e. on SWE. The effect on $r$ is less consistent, improving in three cases but essentially unchanged for SegFormer on SWE (0.111 against 0.113), so joint training improves accuracy consistently while its effect on spatial correlation depends on the target.

To assess the utility of the InSAR features, we also retrained SegFormer removing the five interferometric bands, using only elevation, local incidence angle, the two thermal features and the four temporal encodings. Every metric degrades: MAE rises from 10.391 to 11.082 cm for HS and from 27.113 to 28.143 mm w.e. for SWE, R\textsuperscript{2} drops from $-0.800$ to $-1.320$ and from $-0.667$ to $-1.056$, $r$ falls from 0.065 to 0.042 and from 0.139 to 0.104, the predicted-to-reference standard deviation ratio decreases from 0.54 to 0.36 and from 0.62 to 0.38, and the seed-to-seed spread in MAE grows from 0.445 to 1.723 cm. The gain in absolute error is modest, around 6\% for HS and 4\% for SWE, so much of the predictable signal is carried by elevation and thermal forcing, but the improvement across all four metrics shows that the interferometric observables contribute information the auxiliary inputs do not.

\begin{figure*}
    \centering
    \includegraphics[width=\linewidth]{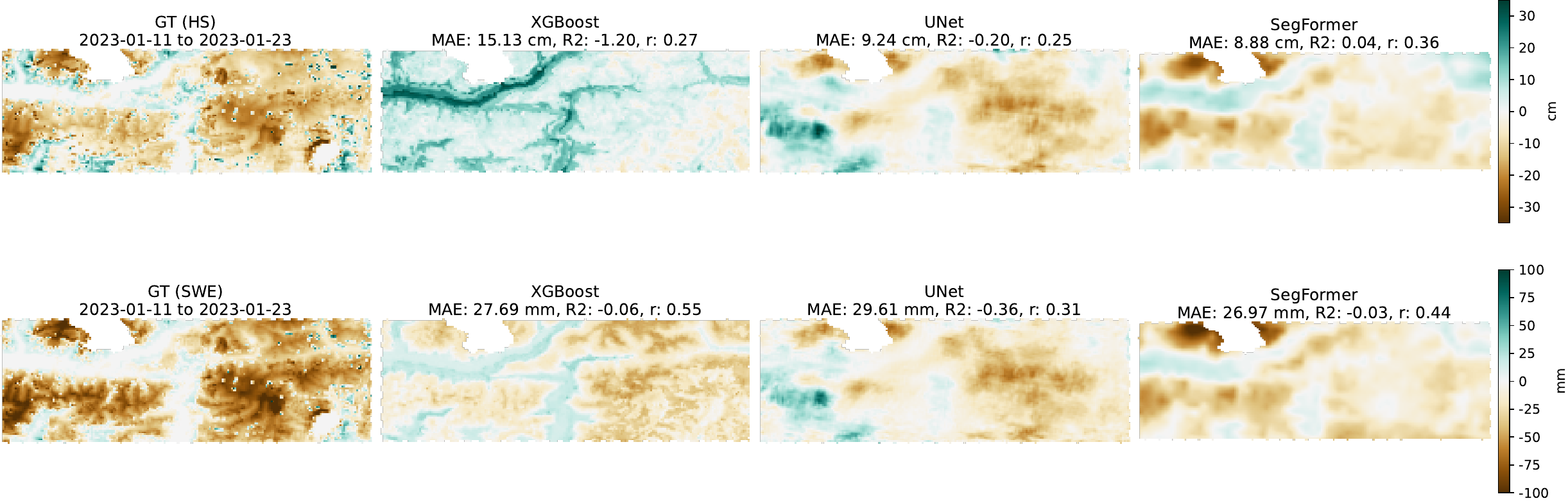} 
    \caption{Qualitative results for the InSAR pair 2023/01/11--2023/01/23, all-features configuration, seed 2024. First row (HS): ground truth, XGBoost, U-Net and SegFormer predictions; second row: the same sequence for SWE. Colour scales are symmetric (blue-green: accumulation, brown: ablation); Metrics refer to this pair only and can differ considerably from the period averages in Table~\ref{tab:single_target}.}
    \label{fig:qualitative}
    
\end{figure*}

Figure~\ref{fig:qualitative} illustrates this behaviour for a single pair (2023/01/11--2023/01/23). On HS, XGBoost yields an MAE of 15.13 cm against 8.88 cm for SegFormer, and its map reproduces the elevation model rather than the snow distribution. SegFormer follows the reference most closely, attaining the only positive R\textsuperscript{2} observed over the test period (0.04) and the highest $r$ (0.36). On SWE, however, XGBoost attains an $r$ of 0.55, above both deep-learning models, with a spatial variability almost entirely explained by the elevation gradient already present in the reference, so a high correlation is obtained without extracting information from the interferometric signal. No single aggregate metric therefore separates the architectures reliably on differential targets: XGBoost attains the highest $r$ of the three models in this window while ranking below both on HS over the test period as a whole.

\section{Conclusions and Future Works}
\label{sec:limitations}
This work considered three ML approaches for estimating SWE and HS variations from aggregated Sentinel-1 InSAR data over the Orobie Alps and Adamello regions. SegFormer attains the lowest error and variability on both targets, and on the all-features baseline the three architectures are separated far more clearly by the spatial metrics than by mean error: R\textsuperscript{2} is $-0.800$ against $-2.138$ for U-Net and $-2.631$ for XGBoost, with Pearson correlations of 0.065, 0.018 and 0.002. Including all features does not guarantee the lowest error for any model, and the informative subset is model- and task-specific. Multi-target training consistently improves accuracy, and removing the interferometric channels degrades every metric.

Several limitations remain. Mean interferometric coherence stays low throughout the test period, between 0.26 and 0.33, which constrains the contribution of the phase-based features; spatial correlation also varies substantially between windows, without a monotonic relationship to positive degree days or mean coherence, so a systematic dependence on thermal regime cannot be established from seven windows. R\textsuperscript{2} is negative for every configuration, and its decomposition attributes 58--73\% of the normalised error to a systematic offset in the window-mean variation, with the remainder from under-dispersion and near-zero pattern correlation; all models under-predict the reference variability. This offset is a target for calibration rather than an intrinsic limit, but our test period does not allow it to be estimated independently. A related confound is that the models are trained on year-round data, including snow-free summer windows, while the test period covers only the accumulation and early ablation season, so the systematic offset cannot be separated from the seasonal composition of the training set. The reference is a reanalysis with no published validation for our test period, and no independent in-situ validation was possible here, since IT-SNOW assimilates over a thousand ground snow-depth sensors without a public list of the contributing networks. Finally, all inputs are aggregated to the 500~m grid from a considerably finer native resolution, averaging out sub-pixel variability in complex terrain. Future work should therefore include temporal cross-validation over longer periods, to validate feature selection on held-out data, and an extension beyond the Italian Alps with a reference independent of the networks used for evaluation. Multi-frequency SAR data, in particular L-band from NISAR, would address the loss of coherence and penetration that limits C-band interferometry in wet snow, which our results suggest is a primary constraint on spatial skill; multi-modal fusion, physics-informed constraints and uncertainty quantification are further avenues towards operational use.

\bibliographystyle{IEEEbib}
\bibliography{strings,refs}

\end{document}